%% file: acl_latex.tex
\pdfoutput=1

\documentclass[11pt]{article}

\usepackage{acl}

\usepackage{times}
\usepackage{latexsym}

\usepackage[T1]{fontenc}

\usepackage[utf8]{inputenc}

\usepackage{microtype}

\usepackage{soul}
\usepackage{url}
\usepackage{breakurl}
\usepackage[breaklinks]{hyperref}

\usepackage[utf8]{inputenc}
\usepackage[font=footnotesize,margin=0.08cm]{caption}
\usepackage{graphicx}
\usepackage{amsmath,amsfonts} 
\usepackage{booktabs}
\usepackage{subcaption}
\usepackage{mathtools}
\usepackage{makecell}

\usepackage{tabu}
\usepackage{stfloats}

\usepackage{enumitem}
\usepackage{siunitx}
\usepackage{multirow}
\usepackage{algorithm}
\usepackage{setspace}
\usepackage{algcompatible}
\usepackage[noend]{algpseudocode}
\makeatletter
\def\BState{\State\hskip-\ALG@thistlm}
\makeatother

\algdef{SE}[SUBALG]{Indent}{EndIndent}{}{\algorithmicend\ }%
\algtext*{Indent}
\algtext*{EndIndent}

\usepackage{xcolor}
\newcommand{\parm}{\mathord{\color{black!33}\bullet}}

\newcommand\METHOD{ROCAAL~}
\newcommand\PRODUCTDT{Product~}
\newcommand\RUMOURDT{Rumour~}
\newcommand\ADRDT{ADR~}
\newcommand\OBSERVATIONDT{Observation~}

\title{Multi-View Active Learning for Short Text Classification in User-Generated Data}

\author{Payam Karisani \\
  Emory University \\
  \texttt{pkarisa@emory.edu} \\\And
  Negin Karisani \\
  Purdue University \\
  \texttt{nkarisan@purdue.edu} \\\And
  Li Xiong \\
  Emory University \\
  \texttt{lxiong@emory.edu} \\}

\begin{document}
\maketitle
\begin{abstract}
    Mining user-generated data often suffers from the lack of enough labeled data, short document lengths, and the informal user language. In this paper, we propose a novel active learning model to overcome these obstacles in the tasks tailored for query phrases--e.g., detecting positive reports of natural disasters. Our model has three novelties: 1) It is the first approach to employ multi-view active learning in this domain. 2) It uses the Parzen-Rosenblatt window method to integrate the representativeness measure into multi-view active learning. 3) It employs a query-by-committee strategy, based on the agreement between predictors, to address the usually noisy language of the documents in this domain. We evaluate our model in four publicly available Twitter datasets with distinctly different applications. We also compare our model with a wide range of baselines including those with multiple classifiers. The experiments testify that our model is highly consistent and outperforms existing models.
\end{abstract}

\input{doc-body}

\section*{Acknowledgements}
We thank the anonymous reviewers for their insightful feedback. The research is partially supported by National Science Foundation under CNS- 2125530. 

\section*{Limitations} \label{sec:ethics}
\input{doc-ethics.tex}

\bibliography{anthology,custom}
\bibliographystyle{acl_natbib}

\appendix

\input{doc-appendix}


\end{document}

%% file: doc-body.tex
\section{Introduction} \label{sec:intro}
\input{doc-intro.tex}

\section{Related Work} \label{sec:rel-work}
\input{doc-related-work.tex}

\section{Proposed Model} \label{sec:method}

\input{doc-method.tex}

\section{Experimental Setup} \label{sec:setup}
\input{doc-experiment-setup.tex}

\section{Results and Analysis} \label{sec:result}
\input{doc-results.tex}

\section{Conclusions} \label{sec:conclusion}

\input{doc-conclusion.tex}

%% file: doc-intro.tex

A microblog is a stream of brief updates written by an author over time on social media platforms such as Twitter or Tumblr \cite{microblogs}. In such platforms, an important set of tasks tailor for queries \cite{co-decomp}. We demonstrate this by an example. Assume we aim to develop a monitoring system for tracking the reports of COVID-19 on the Twitter website. Such a system can be developed in five steps \cite{wespad}: 1)~Collecting a set of general and related query phrases that describe the task. Terms like ``covid-19'', ``covid'', and ``coronavirus'' can constitute such a set. 2)~Collecting the set of tweets that mention these queries. 3)~Sampling a subset of the collected data to be manually annotated. 4)~Training a classifier on the annotated documents. 5)~Using the classifier to label the remaining data.

\begin{figure}
    \centering
    \begin{subfigure}[t]{0.4\linewidth}
        \centering
        \includegraphics[width=\linewidth]{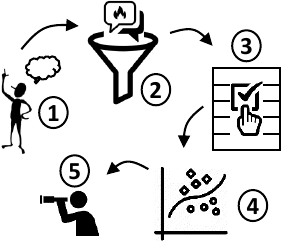}
        \caption{Query-Based Tasks} \label{fig:proc-steps}
    \end{subfigure}~~~
    \begin{subfigure}[t]{0.4\linewidth}
        \centering
        \includegraphics[width=\linewidth]{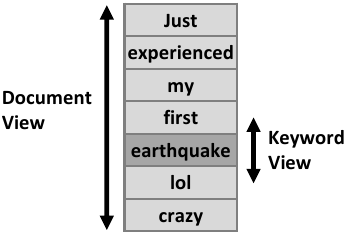}
        \caption{Two Views in Our Model} \label{fig:proc-views}
    \end{subfigure}
    \caption{\textbf{\ref{fig:proc-steps})}~Query-based tasks consist of five steps: 1) obtaining query phrases, 2) collecting the documents that contain the queries, 3) selecting a subset of the collected documents and labeling them, 4) training a classifier, and 5) predicting unseen documents. We use Active Learning to enhance the document selection in the third step and to consequently improve the classifier in the fourth step.  \textbf{\ref{fig:proc-views})}~We extract two views from documents: 1) a document view to encode the entire user posting, and 2) a keyword view to encode the context of the query phrase (earthquake). See Section \ref{subsec:const-multi-view} for more details. }\label{fig:proc}
\end{figure}

A similar pipeline can be used in other tasks
such as detecting suicide ideation \cite{suicide}, monitoring press freedom \cite{press-free}, certain applications for hate speech detection \cite{hate-speech}, certain applications for churn detection \cite{churn}, and general disease detection \cite{our-corona}.
In this study, we employ Active Learning \cite{act-ler-survey} and aim to enhance the document sampling (Step 3) and the classification phase (Step 4) in this pipeline. See Figure \ref{fig:proc-steps} for an illustration.

Existing studies report various difficulties in mining social media data \cite{crisis-survey,active-adr,view-distill}. The costly construction of datasets is a challenge \cite{self-pretraining}. To overcome this challenge, we use Active Learning which is known for reducing the amount of required data by intelligent sampling \cite{act-ler-survey}. Another issue is the typically short length of documents that, along small labeled data, can cause overfitting \cite{view-distill}. To confront this challenge, we aim to use multi-view learning which is known for reducing the risk of overfitting \cite{co-testing-0,cotrain-2} by relying on models trained on different sets of features.\footnote{Multi-view learning is an area of machine learning \cite{multi-view-published} that assumes data points have two or more representations called \textit{views} that can be individually used in algorithms.} See Figure \ref{fig:proc-views} for an illustration of our algorithm for extracting the views. Another challenge is the noisy language\footnote{The word \textit{noise} \cite{nlp-soc-med} refers to the irregular text generated by social media users, it includes but not limited to slang language, misspelled words, out-of-vocabulary words, and ungrammatical sentences. See the series of workshops W-NUT \cite{wnut-2021} on this subject.} of users in social media \cite{wnut-2021}. To address this obstacle, we use an active learning technique based on a committee of predictors. Ensemble learning is known for addressing noisy data by reducing variance \cite{bagging-analysis}. Furthermore, and again related to noisy data, it is reported that social media users are highly inventive in using words \cite{phm-new}. To tackle this challenge, we rely on the context of query phrases to identify word semantics \cite{co-decomp,embed-wsd-survey}.

Our study makes three contributions: 1) Existing active learning models for classifying social media data traditionally rely on single-view algorithms. In this paper, we propose the first multi-view active learning model in this domain. 2) We use pretrained contextual language models to extract our views. To efficiently use these views, we employ the Parzen-Rosenblatt window method \cite{kde-est} and propose a novel query strategy by integrating the representativeness measure into multi-view Active Learning. 3) We use an algorithm based on the agreement between predictors to increase the resistance to social media noise. We empirically demonstrate that this step enhances the selection process in Active Learning.

We evaluate our model, which we term \METHOD (Robust Context-Aware Active Learning), in four distinctly different Twitter tasks and demonstrate that it is either the top model or on a par with the best recent methods.

%% file: doc-related-work.tex
The uncertainty-based sampling model is the most widely used active learning query strategy\footnote{In some papers \cite{nlp-al} the phrase ``acquisition function'' is used instead of the phrase ``query strategy''. However, the word acquisition is traditionally associated with a sub-field of Active Learning called Feature Acquisition. In this paper, we adopt the traditional terminology.} \cite{uncertainty,inactive,active-multi-ans}. In this model, the most \textit{informative} document, i.e., the document that the base learner is uncertain about, is queried and added to the labeled pool. 
\citet{bert-al} survey numerous methods and report that the performance of the classifiers based on pretrained language models can be enhanced by Active Learning, however, they also report that existing query strategies yield no significant gain over the uncertainty-based sampling model--measured by the prediction entropy.

Given the usually satisfactory performance of the uncertainty-based sampling model, the majority of successful applications of Active Learning in microblogging platforms rely on this model \cite{active-adr-2,act-spam,act-job-tweet}. The survey by \citet{al-rumour} confirms this claim, and also reports that language model pretraining can be an additional effective factor to address noisy social media data. Therefore, we follow their suggestion and use pretrained encoders in our model and all of the baselines. 

Incorporating the diversity or the representativeness measures into the uncertainty-based query strategy is an active area of research \cite{cal,caral,alps,al-adv-uncertain}. 
While the uncertainty-based model, which is inherently a single-view model, has shown to be effective, it is well-known that multi-view models have a considerable potential \cite{multi-view}. 

To our knowledge, there is no multi-view active learning model for social media data. In this article, we propose such a model by integrating the representativeness measure into multi-view Active Learning. Multi-view Active Learning \cite{co-testing-0,co-testing-2,sent-cotest} constructs two views (or sets of features) from input data and trains a base learner on each view. Then each base learner is used to label the set of unlabeled data. The data points that are assigned to the opposite classes by the two base learners are detected--these data points are called \textit{\textbf{contention}} points. One data point from this set is annotated and added to the labeled set. In the results and analysis section, we show that our model outperforms existing single-view models as well as the regular multi-view active learning model.

%% file: doc-method.tex
We begin this section by explaining the problem statement, and then, we describe the approach to extract two views from documents. We continue by explaining the query strategy and the technique to tackle social media noise. Finally, we assemble the pieces and provide an overview of the model.

\subsection{Problem Statement} \label{subsec:prob-state}
We are given a small set of labeled user postings $L$, a large set of unlabeled user postings $U$, and the query phrase(s) $Q$, where $Q$ appears in all the documents of $L$ and $U$. We aim to to develop a \textit{selection mechanism} to populate the set $L$ by the documents in $U$. Each document is annotated before being added to the set $L$. The objective is to reduce the error rate of the model trained on the set $L$ in labeling unseen documents.

We propose an active learning method to carry out the document selection step.

\begin{figure}
\centering
\includegraphics[width=2.8in]{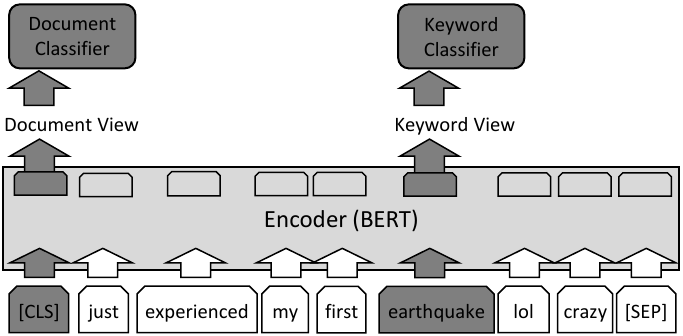}
\caption{Pretrained multi-layer transformers as the encoder in our model. We employ these models--we used BERT in our experiments--to extract the document and the keyword representations. A hypothetical short and simple user posting is shown for demonstration. The token [CLS] in the BERT architecture represents the document encoding. To obtain the keyword representations, we extract the corresponding final pooling layers--see the experimental setup for the details.} \label{fig:two-views}
\end{figure}

\subsection{Extracting Two Contextual Representations from Documents} \label{subsec:const-multi-view}

The approach to construct two views from documents is inspired by the research on Word Sense Disambiguation (WSD) and their mainstream solutions the contextual word embeddings \cite{embed-wsd-survey}. The neural contextual word embeddings are proven to encode the information required to effectively characterize word-level context \cite{co-decomp}. Thus, to extract two contextual representations from documents, we propose to extract one representation on the document level to capture global information about documents, and to extract another representation on the keyword level to capture the context that the query phrases were used in. To extract these representations we use pretrained transformers--e.g., BERT \cite{bert}--as the encoder to construct the document and keyword feature spaces. Since documents always contain at least one of the query phrases, then this task is always feasible. 

We demonstrate this by outlining the task of detecting the true reports of earthquakes on Twitter, see Figure \ref{fig:two-views}. Given the query words ``quake'' and ``earthquake'', we may crawl the hypothetical tweet: \textit{``Just experienced my first \textbf{earthquake} lol crazy''}. Given this tweet, we use the encoder to extract a feature vector on the document level which stores the overall information of the tweet. This corresponds to the tag [CLS] in the BERT architecture, as shown in Figure \ref{fig:two-views}. Additionally, in the same manner we can extract another feature vector on the keyword level to capture the context of the search term,\footnote{In the case that multiple query words are used to collect the data, all the occurrences of the query words in the documents can be mapped to a single synthesized token~\cite{bert-srl}.} i.e., the vector representation of the search term in: \textit{``...my first \textbf{earthquake} lol...''}.

Previous studies \cite{cotrain-ortho,sent-cotest} have shown that correlated views are effective in multi-view learning. Thus, this approach is supported by empirical evidence. In the next section, we exploit these two views in an active learning framework.

\subsection{Integrating Representativeness into Multi-View Active Learning} \label{subsec:exploit-density}

To develop our query strategy, we note that the ranking function, to select the best unlabeled documents, should be proportional to the confidence scores of the base learners in the two views. Because by definition a multi-view model is a contention reduction model, hence, for a document to be informative the classifier outputs must confidently point to the opposite directions. On the other hand, since we desire to incorporate document representativeness, the scores in each view should also represent the concentration of data points in the feature space.

To formally implement this idea, let $\vec{d}_t$ and $\vec{w}_t$ be the document and keyword level representations of the document $t$. The scoring function below follows our desired criteria outlined above:
\begin{equation} \label{eq:score}
\small
\setlength{\jot}{0pt}
\setlength{\abovedisplayskip}{5pt}
\setlength{\belowdisplayskip}{5pt}
\medmuskip=0mu
\thinmuskip=0mu
\thickmuskip=0mu
\nulldelimiterspace=0pt
\scriptspace=0pt
\begin{split}
score(t) =& P_{D}(\vec{d}_t) \times Conf_{D}(\vec{d}_t) 
~+~P_{W}(\vec{w}_t) \times Conf_{W}(\vec{w}_t),
\end{split}
\end{equation}
where $Conf_{D}(\vec{d}_t)$ and $Conf_{W}(\vec{w}_t)$ are the confidence of the classifiers in the document level and in the keyword level views for labeling the example $t$. A classifier confidence is measured by the probability assigned to the output label \cite{cls-conf}. A high probability means a high confidence. Therefore:
\begin{equation} \label{eq:confidence_d}
\small
\setlength{\jot}{0pt}
\setlength{\abovedisplayskip}{5pt}
\setlength{\belowdisplayskip}{5pt}
\medmuskip=0mu
\thinmuskip=0mu
\thickmuskip=0mu
\nulldelimiterspace=0pt
\scriptspace=0pt
\begin{split}
Conf_{D}(\vec{d}_t) = \operatorname*{max}_{y}~~P_{\theta_D}(y|\vec{d}_t),
\end{split}
\end{equation}
and
\begin{equation} \label{eq:confidence_w}
\small
\setlength{\jot}{0pt}
\setlength{\abovedisplayskip}{5pt}
\setlength{\belowdisplayskip}{5pt}
\medmuskip=0mu
\thinmuskip=0mu
\thickmuskip=0mu
\nulldelimiterspace=0pt
\scriptspace=0pt
\begin{split}
Conf_{W}(\vec{w}_t) = \operatorname*{max}_{y}~~P_{\theta_W}(y|\vec{w}_t),
\end{split}
\end{equation}
where the classifiers in the document and in the keyword level views are parameterized by $\theta_D$ and $\theta_W$ respectively. Note that in Equations \ref{eq:confidence_d} and \ref{eq:confidence_w},  $P_{\theta_{\parm}}(y|\parm)$ are real-valued probabilities, so are $Conf_{\parm}(\parm)$. 
In Equation \ref{eq:score}, $P_{D}(\vec{d}_t)$ and $P_{W}(\vec{w}_t)$ are the probabilities of the document $t$ belonging to the set of contention points in the document and in the keyword level views respectively. To obtain these quantities, we use the Parzen-Rosenblatt window method \cite{kde-est}. Therefore, we have:
\begin{equation} \label{eq:prob_d}
\small
\setlength{\jot}{0pt}
\setlength{\abovedisplayskip}{5pt}
\setlength{\belowdisplayskip}{5pt}
\medmuskip=0mu
\thinmuskip=0mu
\thickmuskip=0mu
\nulldelimiterspace=0pt
\scriptspace=0pt
\begin{split}
P_{D}(\vec{d}_t)=\frac{1}{n}~\sum_{i=1}^{n}~\frac{1}{(h_{1})^{b}}~\phi_D (\frac{\vec{d}_t-\vec{d}_{t_i}}{h_1}),
\end{split}
\end{equation}
and
\begin{equation} \label{eq:prob_w}
\small
\setlength{\jot}{0pt}
\setlength{\abovedisplayskip}{5pt}
\setlength{\belowdisplayskip}{5pt}
\medmuskip=0mu
\thinmuskip=0mu
\thickmuskip=0mu
\nulldelimiterspace=0pt
\scriptspace=0pt
\begin{split}
P_{W}(\vec{w}_t)=\frac{1}{n}~\sum_{i=1}^{n}~\frac{1}{(h_{2})^{b}}~\phi_W (\frac{\vec{w}_t-\vec{w}_{t_i}}{h_2}),
\end{split}
\end{equation}
where $n$ is the number of the contention points and $b$ is the number of dimensions--these quantities do not change across the views. $h_1$ and $h_2$ are called bandwidth hyper-parameters. $\phi_D(\parm)$ and $\phi_W(\parm)$ are the kernel functions. $\vec{d}_{t_i}$ and $\vec{w}_{t_i}$ are the representations of the contention document $t_i$ in the document and in the keyword level views respectively. \citet{kde-est} discusses several kernel functions, including the triangular, Gaussian, and Epanechnikov kernels. The triangular kernel is the simplest one, which we use in our model. Thus we have:
\begin{equation} \label{eq:kernel_d}
\small
\setlength{\jot}{0pt}
\setlength{\abovedisplayskip}{5pt}
\setlength{\belowdisplayskip}{5pt}
\medmuskip=0mu
\thinmuskip=0mu
\thickmuskip=0mu
\nulldelimiterspace=0pt
\scriptspace=0pt
\begin{split}
\phi_D(\vec{a})=\left\{\begin{matrix}
1-|\vec{a}| & \text{if} |\vec{a}| < h_1 \\
0 & \text{otherwise}, \\
\end{matrix}\right.
\end{split}
\end{equation}
and
\begin{equation} \label{eq:kernel_w}
\small
\setlength{\jot}{0pt}
\setlength{\abovedisplayskip}{5pt}
\setlength{\belowdisplayskip}{5pt}
\medmuskip=0mu
\thinmuskip=0mu
\thickmuskip=0mu
\nulldelimiterspace=0pt
\scriptspace=0pt
\begin{split}
\phi_W(\vec{a})=\left\{\begin{matrix}
1-|\vec{a}| & \text{if} |\vec{a}| < h_2 \\
0 & \text{otherwise}, \\
\end{matrix}\right.
\end{split}
\end{equation}
where $|\vec{a}|$ is the vector norm.

\begin{figure}
\centering
\includegraphics[width=3.0in]{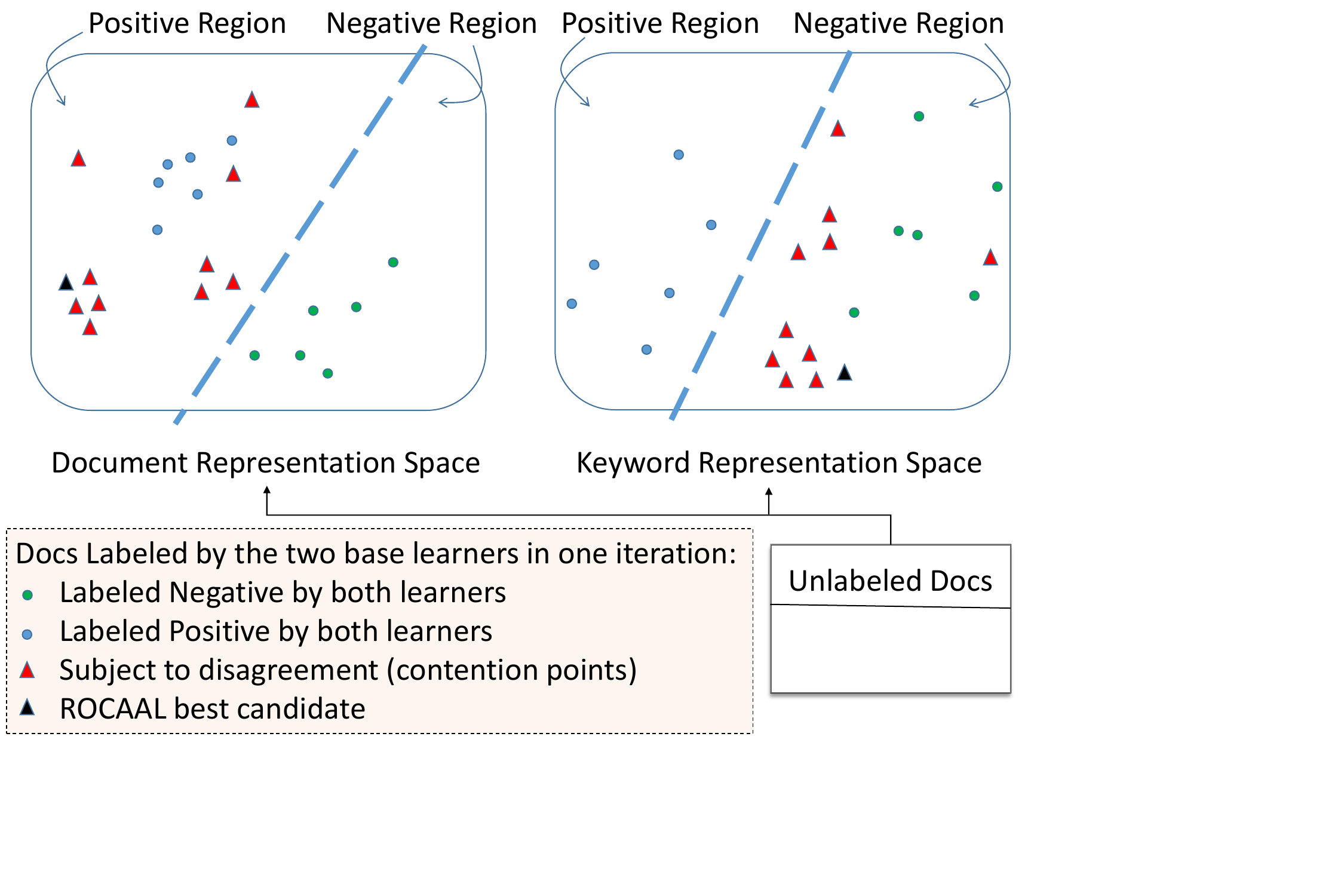}
\caption{The document and keyword level views. \METHOD queries the contention point which is closest to the set of other contention points and also has a large distance from the decision boundaries in the two views (the black triangle). Figure best viewed in color.} \label{fig:qs}
\end{figure}

Intuitively, our scoring function (Equation \ref{eq:score}) assigns a higher score to the documents that are confidently assigned to the opposite classes in the two views (i.e., have a large distance from the decision boundaries in the two views), and are also close to the other set of contention points in each view. Figure \ref{fig:qs} illustrates the principle. Each data point in the document representation space (the left panel) is associated to one data point in the keyword representation space (the right panel). The triangular data points are the set of contention documents, i.e., the documents that are assigned to the opposite classes by the classifiers in the two views. Based on our query strategy, the best candidate document has three properties: 1) It belongs to the set of contention data points, i.e., the triangular data points. 2) It is subject to the most disagreement between the two base learners. That is, it has a large distance from the decision boundaries, this is quantified by $Conf_{\parm}(\parm)$ in Equation \ref{eq:score}. 3) It is close to the other set of contention data points. This is quantified by $P_{\parm}(\parm)$ in Equation \ref{eq:score}.

The document that possesses the above three properties is indicated by a black triangle in Figure~\ref{fig:qs}. We see that by querying this document, we can effectively infer the labels of the patch of red data points close to this document in both views.

There are two advantages in employing this scoring function. First, scaling the confidence of the base learners by the probability densities naturally aggregates the benefits of the contention reduction \cite{qbc-by-bag} and the density based \cite{density-clustering} query strategies. Second, assuming that the points that are close to each other in the feature space are similar and are likely to have the same label,\footnote{This can be justified by the cluster hypothesis \cite{cluster-kernel}.} by promoting the documents that are close to the cluster of other contention points, we can effectively use the contextual information to resolve the disagreement over a set of similar documents. This is particularly the case when a candidate document and its adjacent points are projected into the same regions in both views. 


\subsection{Enhancing Resistance to Noise} \label{subsec:incorp-bag}
As pointed out by \citet{nlp-soc-med}, the documents in social media--particularly on the Twitter website--are highly noisy. They tend to be short, and are packed with inventive lexicons. For instance, in our early example of extracting the reports of earthquakes, a document may be added to the set of contention points and selected for annotation by accident due its noisy content, e.g., the existence of irregular or figurative language.\footnote{For example, using the word earthquake as a reference to excitement. See \citet{figurative-survey} for more information on such linguistic irregularities on social media.} However, selecting another document for annotation might be a better choice to have a diverse and representative training set. If we assume relatively uninformative documents are noise--which due to their unique characteristics may receive a high score by Equation \ref{eq:score}--then we may be able to dampen their effect by variance reduction algorithms.

To address this issue we propose to employ multiple predictors. Bagging is empirically shown to reduce model variance \cite{bagging-analysis}. In the discussed example, bagging can influence the score of the mentioned document, either through affecting the distribution of the contention documents, or reducing the disagreement rate between the two base learners. While it is well-known that bagging is effective for model prediction \cite{intro-ml-book}, we haven't found any recent study to further investigate the utility of bagging for decision making in Active Learning \cite{qbc-by-bag}. In the analysis section, we empirically demonstrate that our proposed technique (see below) for bagging not only improves model prediction, but also enhances the decision making by promoting better candidate documents. We also particularly show that this step enables our model to outperform the baselines that use the regular bagging and also those that use multiple classifiers.

We use this technique as follows: In each iteration, we sample multiple subsets of documents from the set of labeled data. On each subset, we train a pair of base learners as described in Section \ref{subsec:const-multi-view}. For each pair of base learners, we use the model described in Section \ref{subsec:exploit-density} to assign a score to all the unlabeled documents. Finally, the ultimate ranking list is constructed by aggregating the scores of the unlabeled data across the models.

Our technique is different from the regular bagging model \cite{qbc-by-bag}. In the regular bagging model, one estimator is trained on each subset of data, and the best candidate data point is the data point which is subject to the most \textit{disagreement} among the estimators. In our model, the candidate data points, for each subset, are the data points that are assigned to the opposite classes by the base learners. Then, each pair of base learners vote for the candidate data points, and the best candidate data point is the one that is subject to the most \textit{agreement} among all the pairs.

\begin{algorithm}[ht]
\small
\caption{Single Iteration of \METHOD}\label{alg:summary}
\begin{algorithmic}[1]
\algrenewcommand\algorithmicindent{7pt}
\Procedure{\METHOD}{}
    \State \textbf{Given:}
    \Indent
        \State $L: \text{Set of labeled documents}$ 
        \State $U: \text{Set of unlabeled documents}$
        \State $T: \text{Set of test documents}$
        \State $K: \text{Number of estimators for bagging}$
    \EndIndent
    \State \textbf{Return:}
    \Indent
        \State $\text{Labeled set of test documents \& updated training set}$
    \EndIndent
    \State \textbf{Execute:}
    \State Define $S$ as 1-d array // $\text{to store the sub-samples}$
    \State Define $BL$ as 2-d array // $\text{to store the base learners}$
    \State Define $DS$ as 2-d array // $\text{to store the Parzen models}$
    \State Define $C$ as 1-d array // $\text{to store the contention docs}$
    \State Define $Conf$ as 2-d array // $\text{to store confidence scores}$
    \State Define $P$ as 2-d array // $\text{to store the probability values}$
    
    \Indent
        \For{$i \gets 1$ to $K$} \label{alg-line:train-begin}
            \State Sample a subset of L and store in $S[i]$
            \State Train \parbox[t]{.8\linewidth}{two base learners on the two views of $S[i]$ and store in $BL[i][0]$ and $BL[i][1]$}
            \State Use $BL[i][0]$ and $BL[i][1]$ to label the set $U$
            \State Store \parbox[t]{.8\linewidth}{the contention documents in $C[i]$, and their prediction confidences in $Conf[i][0]$ and $Conf[i][1]$}
            \State Train \parbox[t]{.8\linewidth}{two Parzen models on the two views of $C[i]$ and store them in $DS[i][0]$ and $DS[i][1]$}
            \State Use \parbox[t]{.8\linewidth}{$DS[i][0]$ and $DS[i][1]$ to calculate the probability mass values for all the documents in $C[i]$ and store them in $P[i][0]$ and $P[i][1]$}
        \EndFor
        \State Plug \parbox[t]{.8\linewidth}{the arrays $Conf$ and $P$ into Equation (\ref{eq:score}) to calculate the aggregated score for documents in $C$} \label{alg-line:aggregate}
        \State Rank \parbox[t]{.8\linewidth}{all the documents in $C$ based on their score, and store the top one in the new variable $W$}
        \State Query the label of $W$ \label{alg-line:query} // \textbf{ Active Learning Query}
        \State Add $W$ to $L$ and to every $S[\parm]$ in $S$ \label{alg-line:add}
        \State Retrain \parbox[t]{.8\linewidth}{the base learners in $BL[\parm]$ on their corresponding updated sets in $S$} \label{alg-line:train-end}
        \For{$doc$ in $T$}  \label{alg-line:test-begin}
            \State $PCount \gets 0$
            \For{$cls\_pair$ in $BL$}
                \State Use \parbox[t]{.75\linewidth}{the two classifiers in $cls\_pair$ to label the document $doc$ and aggregate the outputs, if the final label is positive then increment $PCount$} \label{alg-line:labeling}
            \EndFor
            \State If \parbox[t]{.8\linewidth}{$(PCount \geq K / 2)$ then $doc$ is positive, otherwise it is negative} \label{alg-line:maj}
        \EndFor \label{alg-line:test-end}
        \State $\text{Return }T, L$
    \EndIndent
\EndProcedure
\end{algorithmic}
\end{algorithm}

\subsection{Overview of Algorithm} \label{subsec:alg}
Algorithm \ref{alg:summary} summarizes \textit{one iteration} of \METHOD\!. Here we leave out the implementation details and only mention the primary steps.

Lines \ref{alg-line:train-begin}-\ref{alg-line:train-end} describe the training procedure, and Lines \ref{alg-line:test-begin}-\ref{alg-line:test-end} describe the labeling procedure. The training stage begins by sampling from the set of labeled documents; then two base learners are trained on the two views of the sampled set. The two base learners are used to label the set of unlabeled documents. The contention documents are detected, and in each view one Parzen-Rosenblatt window method is trained. The two models are used to approximate the probability mass values of every contention document. These steps are repeated for each sub-sample. To rank the set of unlabeled documents, the prediction confidences and probability mass values are used in Equation \ref{eq:score} to score all of the contention documents. One document with the highest score is selected and queried to be added to the labeled set and all of the sampled sets--Line \ref{alg-line:query} and Line \ref{alg-line:add}. Finally, all of the base learners are re-trained on the updated sampled sets. In the labeling stage, each pair of the base learners is used to label the test documents--Line \ref{alg-line:labeling}. To predict the final label, a majority voting algorithm is used--Line \ref{alg-line:maj}.

%% file: doc-experiment-setup.tex

In this section we discuss the datasets, the baselines, and the implementation details. See Appendix \ref{sec:append-data-anal} for more details on the used datasets.

\subsection{Datasets} \label{subsec:dataset}

Social media is a multi-million dollar industry. It can be weaponized to target the basis of democracy or it can be a powerful tool for humanitarian aid. However, as stated in Section \ref{sec:intro}, mining this resource is challenging, which is the motivation for our research. To extensively evaluate our model we use four distinctly different and publicly available Twitter datasets.

\noindent\textbf{Detecting reports of product consumption.} We use the dataset introduced by \citet{smm4h} for an ACL shared task. A document in this dataset is positive if it reports consuming a product. We use the product references to construct the keyword view.

\noindent\textbf{Detecting rumours.} We include the dataset introduced by \citet{rumour-dt}, the revision prepared by \cite{dom-ada-mixture-bert}. In this dataset a document is positive if it spreads a rumour. We use the rumour phrases to construct the keyword view.

\noindent\textbf{Detecting reports of medical drug side-effects.} We use the dataset introduced and expanded by \citet{smm4h-2021}. This dataset has been used for several consecutive years in various NLP shared tasks including NAACL 2021. A document in this dataset is positive if it reports the side-effects of a drug. We use the drug references to construct the keyword view.

\noindent\textbf{Detecting reports of observations.} We include the dataset introduced by \citet{crisis-wit}. In this dataset a document is positive if it reports direct experience with a natural crisis, e.g., an earthquake. We use the crisis phrases to construct the keyword view.

Table \ref{tbl:dt-stat} reports the size of each dataset, along the number of positive and negative documents in each one. We see that the rumour dataset is the smallest one and the ADR dataset is the largest one, with ADR being the most imbalanced benchmark.

\begin{table}
\centering
\small
\begin{tabu}{p{1in} p{0.45in} p{0.35in} p{0.35in} p{0.35in} } \Xhline{3\arrayrulewidth}
\multicolumn{1}{c}{\textbf{Dataset}} & \textbf{Set} & \textbf{\# Doc} & \textbf{\# Neg} & \textbf{\# Pos} \\ \Xhline{3\arrayrulewidth}
\multicolumn{1}{c}{\multirow{4}{*}{\PRODUCTDT}} & Train & 4,503 & 3,104 & 1,399 \\
\multicolumn{1}{c}{} & Test & 2,114 & 1,648 & 466 \\ 
\cmidrule(l){2-5}
\multicolumn{1}{c}{} & Total & 6,617 & 4,752 & 1,865 \\ \Xhline{3\arrayrulewidth}
\multicolumn{1}{c}{\multirow{4}{*}{\RUMOURDT}} & Train & 4,001 & 2,641 & 1,360 \\
\multicolumn{1}{c}{} & Test & 1,801 & 1,189 & 612 \\ 
\cmidrule(l){2-5}
\multicolumn{1}{c}{} & Total & 5,802 & 3,830 & 1,972 \\ \Xhline{3\arrayrulewidth}
\multicolumn{1}{c}{\multirow{4}{*}{\ADRDT}} & Train & 20,624 & 18,659 & 1,965 \\
\multicolumn{1}{c}{} & Test & 4,992 & 4,581 & 411 \\ 
\cmidrule(l){2-5}
\multicolumn{1}{c}{} & Total & 25,616 & 23,240 & 2,376 \\ \Xhline{3\arrayrulewidth}
\multicolumn{1}{c}{\multirow{4}{*}{\OBSERVATIONDT}} & Train & 7,998 & 5,694 & 2,304 \\
\multicolumn{1}{c}{} & Test & 6,001 & 4,911 & 1,090 \\ 
\cmidrule(l){2-5}
\multicolumn{1}{c}{} & Total & 13,999 & 10,605 & 3,394 \\ \Xhline{3\arrayrulewidth}
\end{tabu}
\caption{The size and the class distribution of the datasets.} \label{tbl:dt-stat}
\end{table}

\subsection{Baselines} \label{subsec:baselines}
We compare our model with a wide range of baselines, including those that use bagging and use multiple classifiers. All the models (the baselines and our model) use pretrained BERT as the encoder--see the next section for details.

\noindent\textbf{random:} This baseline is without Active Learning. In each iteration, we randomly select one document from the set of unlabeled data and add to the labeled set. Then, retrain the base learner.

\noindent\textbf{uncertainty:} It is the uncertainty-based model \cite{act-ler-survey}. The output entropy of the base learner was used as the selection criterion.

\noindent\textbf{qbc:} It is a query-by-committee model constructed via bagging with 20 classifiers and 60\% sampling ratio \cite{act-ler-survey}. In this model, the document with the highest rate of disagreement between the classifiers is selected for annotation.

\noindent\textbf{lal:} It is an ensemble with 20 predictors proposed by \citet{lal}. This model is a meta-learning algorithm. The authors argue that instead of manually crafting query strategies, a model should be able to learn the query strategy. They propose to use a regressor that learns the best strategy in the given task. The model estimates the expected error of every data point, then queries the data point that maximizes the error reduction.

\noindent\textbf{caral:} It is proposed by \citet{caral}. In this model, the informativeness of data points is defined by the variation in their consecutive predictions during the training of the classifier. The model relies on data maps \cite{data-maps}, which uses these predictions to categorize data points into easy, ambiguous, and hard. The authors argue that the data points on the boundary of ambiguous and hard categories are the best candidates and contain the highest diversity.

\noindent\textbf{cal:} It is proposed by \citet{cal}. In this model, the diversity and informativeness are combined by choosing the data point that is most similar to its neighbours, but is assigned to the opposite class labels. To contrast two data points, the authors use the Kullback-Leibler divergence \cite{kl-div} between the classifier predictions.

\subsection{Implementation Details} \label{subsec:implement-detail}

We adopt the standard practice in the active learning literature to carry out the experiments \cite{act-ler-survey,active-practical}. In the cold start state, we randomly sample 50 labeled documents, and assume that the rest of the labeled data is unlabeled. This is a standard practice in the literature \cite{act-ler-survey}; active learning methods enhance the quality of selected data as the algorithm proceeds \cite{active-practical,nlp-al}. We report performance in the test set as the training set is augmented with new labeled documents. Following the argument made by \citet{perf-metric} about imbalanced datasets, to consider the classifier quality (Precision) and coverage (Recall) we report the F1 in the minority set. We repeat all the experiments 5 times with different random seeds and report the average of the runs. 

We use pretrained BERT base \cite{bert,bert-impl} as the encoder in all the models.\footnote{We pretrain the publicly available BERT base variant on the set of unlabeled in-domain documents for each task. These models were used in all the baselines and in our model. Thus, the setting is completely fair.} Note that this makes any improvement difficult, because the pretrained transformers are already robust in data scarce settings \cite{bert}, and any improvement should be additive to these baselines.\footnote{In order to account for the increasing size of the training sets during the active learning iterations, every 350 iterations we fine-tune BERT and update the entire set of document and word representations in our model and in all the baselines.} The size of the vectors in BERT is 768 and as suggested by \citet{bert} we use the average of the last 4 layers to create the vectors--they suggest this based on empirical evidence. We use a one-layer fully connected network as the classifier in all the models. Thus, all the models use an identical network and an identical pretraining/fine-tuning procedure, therefore, their comparison is completely fair.

To implement our model, when multiple mentions of the same keyword are included in the same document, the representation of the first one is used. If the queries consist of multi-word phrases, for simplicity, we use the first word in the sequence as the keyword representation. However, an alternative is to replace the phrases with a synthesized token \cite{co-decomp,bert-srl}. In Equations \ref{eq:prob_d} and \ref{eq:prob_w}, the variable $b$ is the size of the BERT vectors, that is 768. There are multiple ways to set the bandwidths $h_1$ and $h_2$ in the Parzen-Rosenblat window method \cite{kernel-de}. We set these quantities to the average distance of the data points in the document and in the keyword level views respectively (30 and 45), which is independent of labeled data. In our bagging algorithm, we use 10 estimators with 60\% sampling ratio.

%% file: doc-results.tex
In this section we report the results and then we provide an empirical analysis.

\subsection{Results} \label{subsec:result-main}

\begin{figure*}
\centering
        \begin{subfigure}[tb]{0.22\linewidth}
            \centering
            \includegraphics[width=\linewidth]{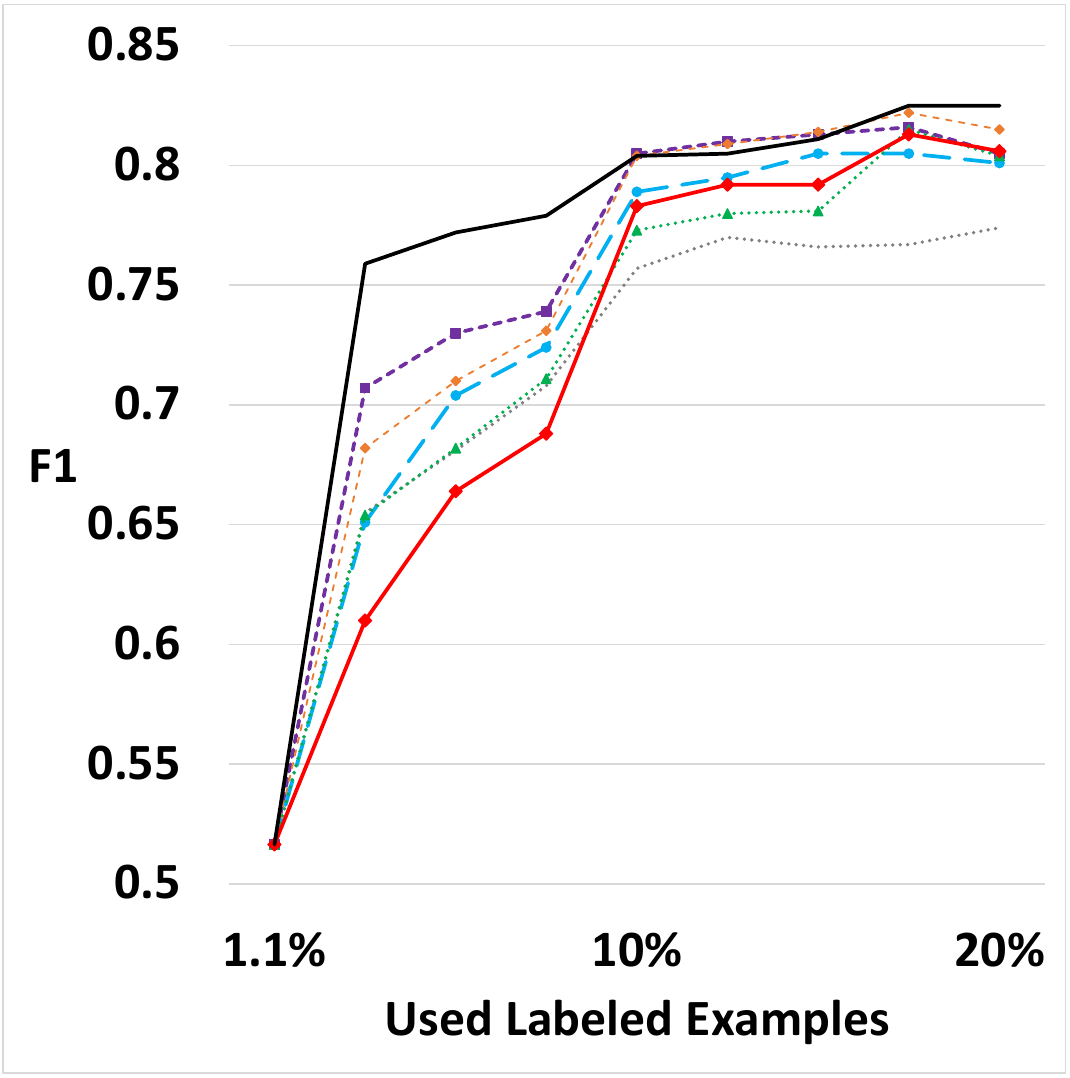}
            \caption{\PRODUCTDT}
            \label{fig:curve-product}
        \end{subfigure}~
        \begin{subfigure}[tb]{0.22\linewidth}
            \centering
            \includegraphics[width=\linewidth]{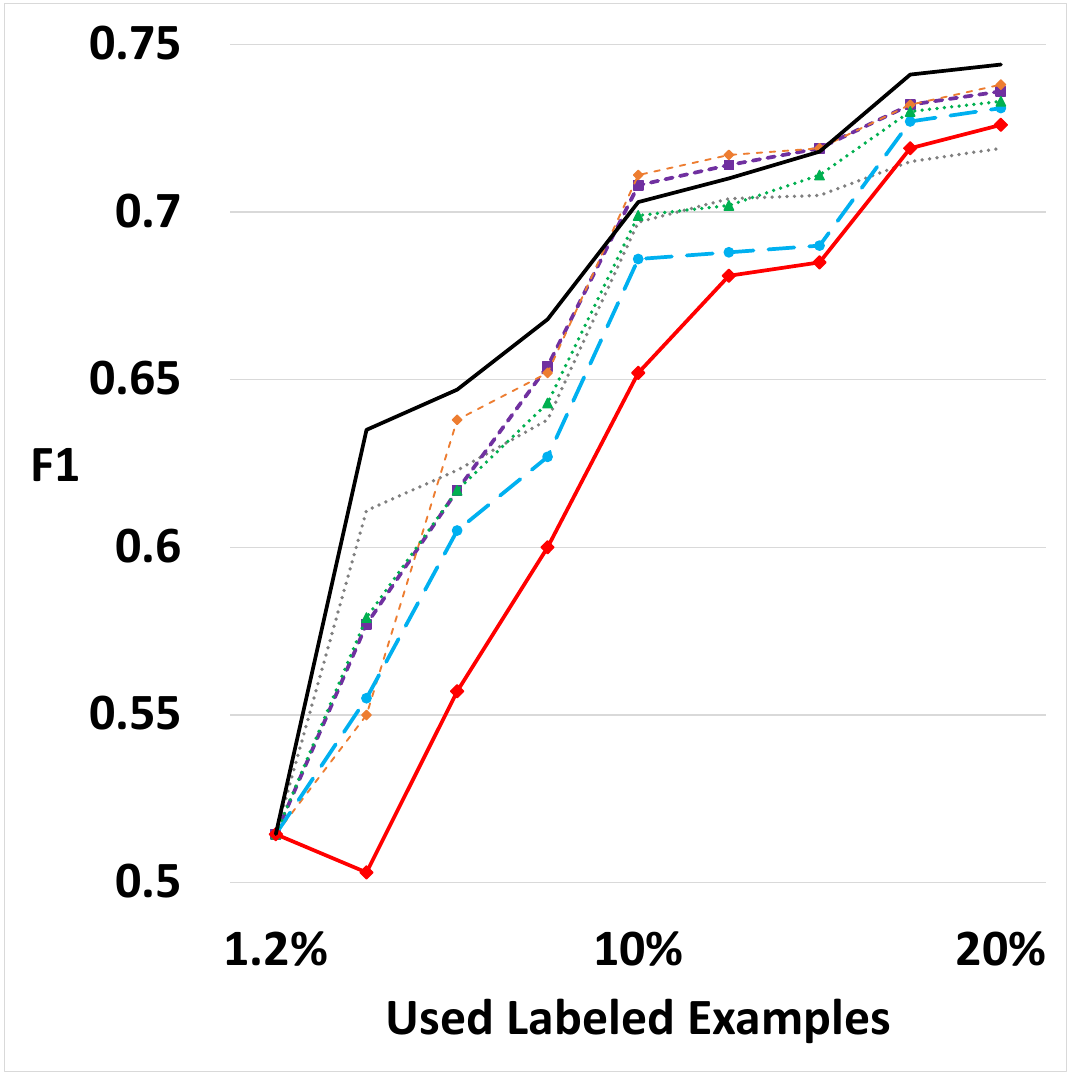}
            \caption{\RUMOURDT}
            \label{fig:curve-rumour}
        \end{subfigure}~
        \begin{subfigure}[tb]{0.22\linewidth}
            \centering
            \includegraphics[width=\linewidth]{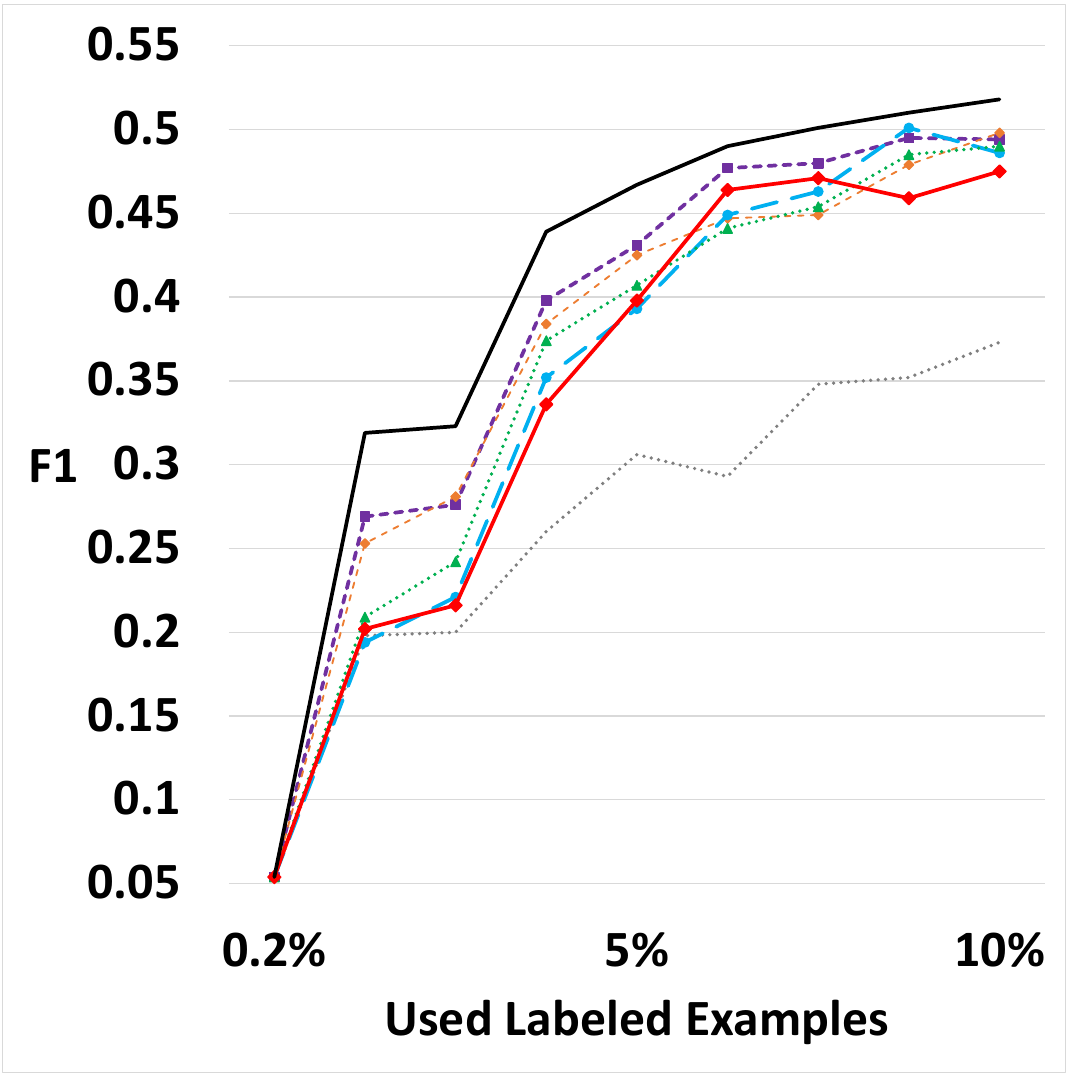}
            \caption{\ADRDT}
            \label{fig:curve-adr}
        \end{subfigure}~
        \begin{subfigure}[tb]{0.22\linewidth}
            \centering
            \includegraphics[width=\linewidth]{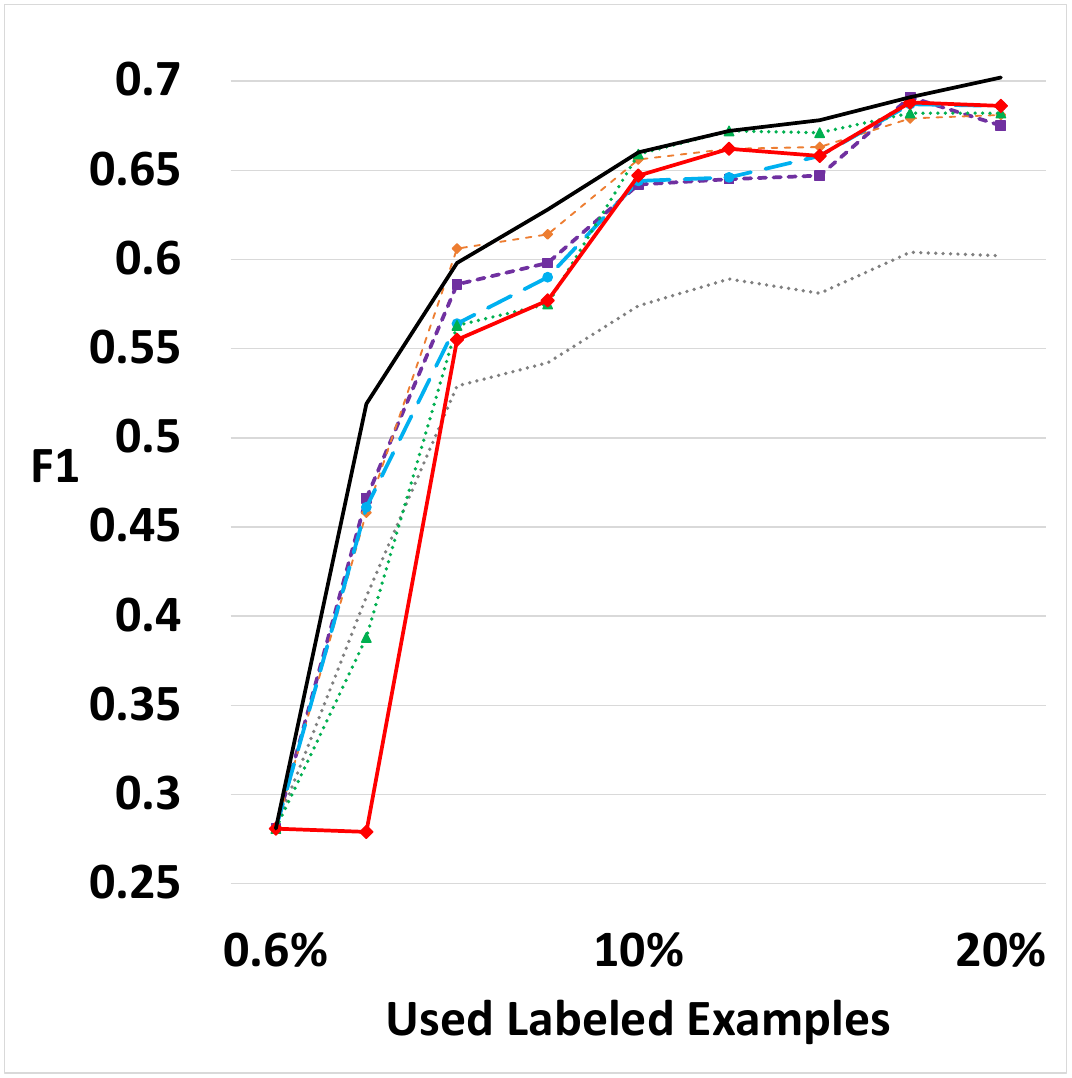}
            \caption{\OBSERVATIONDT}
            \label{fig:curve-observation}
        \end{subfigure}\hfill
        \begin{subfigure}[tb]{0.5\textwidth}
            \centering
            \includegraphics[width=\linewidth]{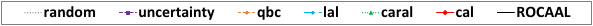}
        \end{subfigure}
        \caption{Learning curves of the models in the four datasets. Note that the horizontal axis units are 5 percent absolute value. We see that as more labeled data is used all the models almost converge \cite{inactive}. Figure best viewed in color.}\label{fig:curve-all}
\end{figure*}

Figure \ref{fig:curve-all} reports the performance of our model compared to that of the baselines in all the datasets. The results confirm that--except in a few cases--all the models outperform \textit{random} baseline, confirming that Active Learning is effective in these tasks. The experiments also show that \textit{uncertainty} model is performing very well, confirming the consistency of this model which is discussed in Section \ref{sec:rel-work} and is also reported in other studies \cite{inactive,bert-al}. Finally, the results signify that our model \METHOD consistently improves over the baselines. During the early iterations, our model exploits two views to issue the queries, whereas the other models rely on one view. This is significant, since in real-world scenarios the set of labeled data points is small and costly to obtain. As more training data becomes available and the pool of unlabeled data shrinks, the models converge \cite{inactive}. Nonetheless, our model still maintains a noticeable superiority. Table \ref{tbl:result-f1} reports the final performance of the models. 

\begin{table*}
\centering
\small
\begin{tabu}{p{0.30in} p{0.75in} p{0.75in} p{0.75in} p{0.75in} } \Xhline{2\arrayrulewidth}
 \cline{1-5} & \multicolumn{4}{c}{\textbf{F1 in datasets}} \\
\cmidrule[\heavyrulewidth](l){2-5} 
\multicolumn{1}{c}{\textbf{Method}} & 
\textbf{Product} & \textbf{Rumour} & \textbf{ADR} & \textbf{Observation} \\ \Xhline{3\arrayrulewidth}
\multicolumn{1}{c}{\textit{random}} & 0.774$\pm$0.01 & 0.719$\pm$0.02 & 0.373$\pm$0.06 & 0.602$\pm$0.02 \\
\multicolumn{1}{c}{\textit{uncertainty}} & 0.805$\pm$0.01 & 0.736$\pm$0.01 & 0.494$\pm$0.02 & 0.675$\pm$0.01 \\
\multicolumn{1}{c}{\textit{qbc}} & 0.815$\pm$0.01 & 0.738$\pm$0.02 & 0.498$\pm$0.02 & 0.681$\pm$0.01 \\
\multicolumn{1}{c}{\textit{lal}} & 0.801$\pm$0.01 & 0.731$\pm$0.00 & 0.486$\pm$0.03 & 0.686$\pm$0.02 \\ 
\multicolumn{1}{c}{\textit{caral}} & 0.804$\pm$0.00 & 0.733$\pm$0.01 & 0.490$\pm$0.01 & 0.682$\pm$0.01 \\ 
\multicolumn{1}{c}{\textit{cal}} & 0.806$\pm$0.01 & 0.726$\pm$0.02 & 0.475$\pm$0.03 & 0.686$\pm$0.02 \\  \hline 
\multicolumn{1}{c}{\textit{\METHOD}} & \textbf{0.825$\pm$0.01} & \textbf{0.744$\pm$0.01} & \textbf{0.518$\pm$0.02} & \textbf{0.702$\pm$0.01} \\ \Xhline{3\arrayrulewidth}
\end{tabu}
\caption{Final F1 measure of our model compared to that of the baselines in all the datasets.} \label{tbl:result-f1}
\end{table*}

\subsection{Empirical Analysis} \label{subsec:analysis}
We begin this section by reporting an ablation study on the efficacy of the individual views. Then, we report a second ablation study on the efficacy of each module in our model (i.e., our query strategy and our variance reduction technique) and also compare with the regular multi-view active learning. Then we evaluate the impact of our variance reduction technique on the decision making in active learning. Finally, we discuss the resource complexity and the hyper-parameter sensitivity of our model.

\begin{table}
\centering
\small
\begin{tabu}{p{1.0in}  p{0.35in} p{0.45in} p{0.45in} } \Xhline{3\arrayrulewidth}
\textbf{Method} & \textbf{F1} & \textbf{Precision} & \textbf{Recall} \\ \Xhline{3\arrayrulewidth}
Keyword view & 0.452 & 0.528 & 0.398 \\ 
Document view & 0.494 & 0.639 & 0.406 \\ 
\METHOD & 0.518 & 0.606 & 0.454 \\ \Xhline{3\arrayrulewidth}
\end{tabu}
\caption{The performance in the document level and keyword level views compared to \METHOD.} \label{tbl:views}
\end{table}

To demonstrate the efficacy of the views proposed in Section \ref{subsec:const-multi-view}, we report an ablation study by leaving out each view and training a model on the remaining view. Table \ref{tbl:views} reports the results. We see that both views have a contribution. We particularly see that the model trained on the document view has a much better performance. This experiment and the next ones require to run a model numerous times. We carried them out in \ADRDT dataset.

We report a second ablation study on the role of our novel active learning query strategy (Section \ref{subsec:exploit-density}) and on the role of our technique for tackling the effect of noise in informal social media documents using a variance reduction method (Section \ref{subsec:incorp-bag}). In each case, the new model is obtained by leaving out one module and evaluating the remaining module. Table \ref{tbl:ablation} reports the results of this experiment. We see that both modules have noticeable influence. In this experiment, we also compare our model with the regular multi-view active learning. This model is obtained by deactivating both modules. We see that it is markedly outperformed.

To demonstrate that our variance reduction technique (Section \ref{subsec:incorp-bag}) specifically improves the candidate selection in Active Learning we need to disentangle the impact of this algorithm from the prediction. To this end, we run two variants of our model. In the first variant we use multiple predictors in the query strategy to select the best candidate document, then we randomly select one pair of the predictors for labeling. In the second variant we use one pair of predictors in the query strategy, then we create a pool of multiple predictors for labeling. Table \ref{tbl:decition-making} reports the results of this experiment. We observe that our algorithm improves both the query selection and the prediction (in terms of F1).\footnote{We don't change the classification threshold here. See Lines \ref{alg-line:test-begin}-\ref{alg-line:maj} in Algorithm \ref{alg:summary} for the labeling procedure.} We see that there is an increase in the precision when we use multiple estimators in the query strategy.

\begin{figure}
    \centering
    \begin{subfigure}[t]{0.45\linewidth}
        \centering
        \includegraphics[width=\linewidth]{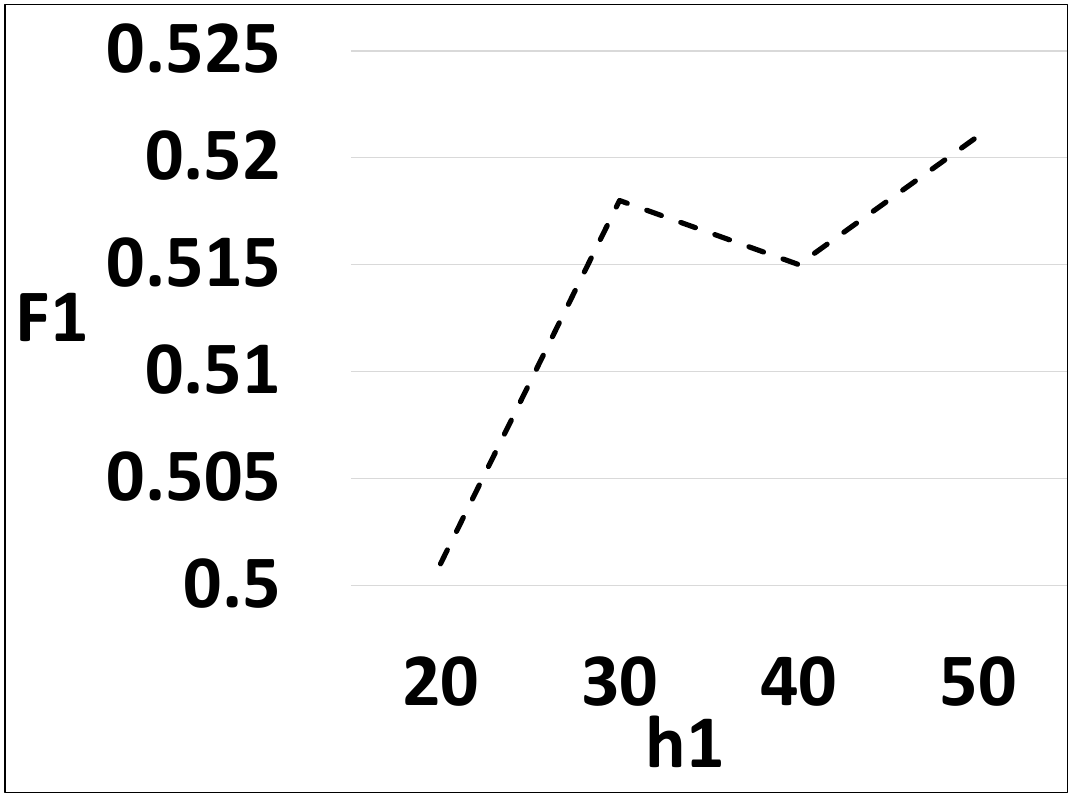} \label{fig:h1}
        \caption{F1 vs h1}
    \end{subfigure}~~~~
    \begin{subfigure}[t]{0.45\linewidth}
        \centering
        \includegraphics[width=\linewidth]{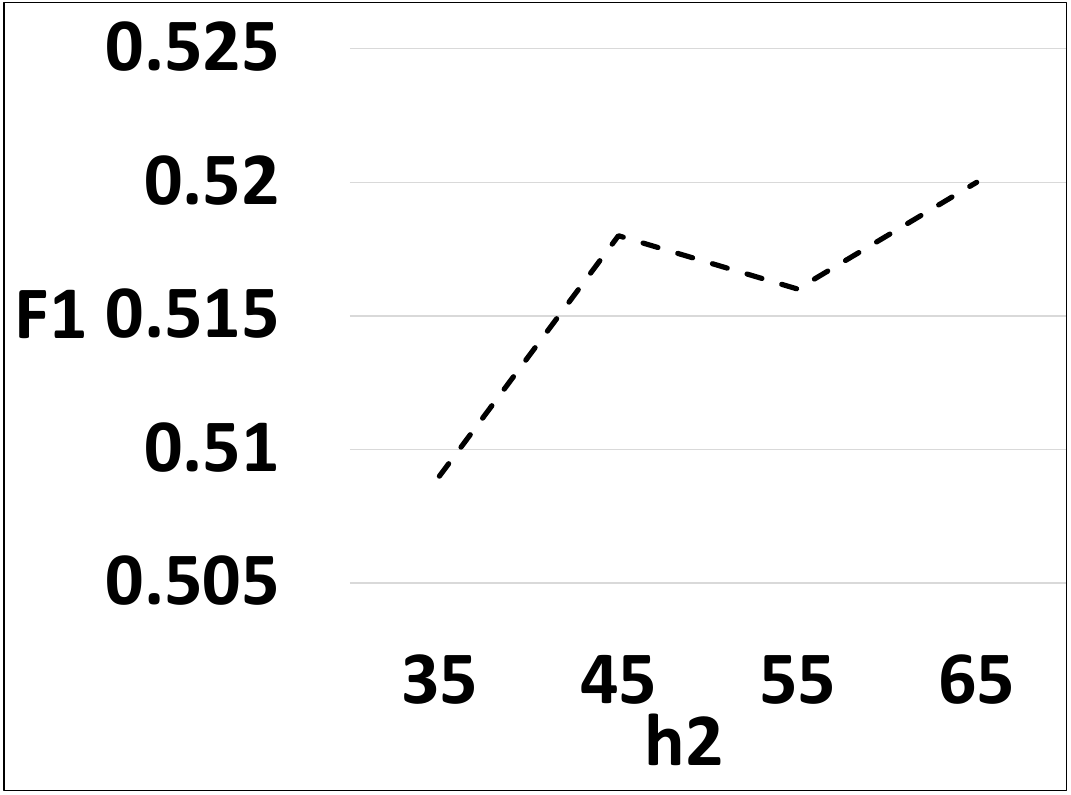} \label{fig:h2}
        \caption{F1 vs h2}
    \end{subfigure}
    \caption{The sensitivity of \METHOD to the hyper-parameters. }\label{fig:h1-and-h2}
\end{figure}

In terms of runtime, our model takes about 10 seconds on average to make one query in \ADRDT dataset--using a system with a 16-core processor and an NVIDIA Titan RTX GPU. 
One particularly interesting quality of our model is the absence of critical hyper-parameters to tune. Excluding the hyper-parameters of the base learners, which is shared between all the baselines, in our experiments \METHOD was not sensitive to other hyper-parameters. Figure \ref{fig:h1-and-h2} reports the performance of our model at varying values of the bandwidths $h1$ and $h2$. We set these quantities to 30 and 45 based on the average distance of the data points in the document and the keyword level views, which is independent of the labeled data. We see that the performance reaches a plateau after a certain threshold. We used \{10,15,20\} as the number of estimators and used \{0.6,0.7,0.8\} as the sampling ratio in bagging, but didn't observe a noticeable sensitivity.

In summary, we used four publicly available datasets in the experiments, and also included multiple state-of-the-art and traditional baselines, and showed that our model consistently outperforms them. We also reported detailed experiments and empirically analyzed our model from multiple aspects. 

\begingroup
\setlength{\tabcolsep}{3pt} 
\begin{table}
\centering
\small
\begin{tabu}{p{1.6in}  p{0.3in} p{0.45in} p{0.35in} } \Xhline{3\arrayrulewidth}
\textbf{Method} & \textbf{F1} & \textbf{Precision} & \textbf{Recall} \\ \Xhline{3\arrayrulewidth}
Regular multi-view & 0.496 & 0.618 & 0.415 \\ 
\METHOD (only varian. reduct.) & 0.507 & 0.613 & 0.433 \\ 
\METHOD (only query strategy) & 0.508 & 0.650 & 0.416 \\ 
\METHOD & 0.518 & 0.606 & 0.454 \\ \Xhline{3\arrayrulewidth}
\end{tabu}
\caption{Ablation study on the efficacy of the query strategy and the variance reduction technique. We see that both steps are equally contributing.} \label{tbl:ablation}
\end{table}
\endgroup
\begin{table}
\centering
\small
\begin{tabu}{p{1.1in}  p{0.3in} p{0.40in} p{0.40in} } \Xhline{3\arrayrulewidth}
\textbf{Method} & \textbf{F1} & \textbf{Precision} & \textbf{Recall} \\ \Xhline{3\arrayrulewidth}
Bagging for selection & 0.512 & 0.650 & 0.423 \\ 
Bagging for prediction & 0.513 & 0.638 & 0.430 \\ 
\METHOD & 0.518 & 0.606 & 0.454 \\ \Xhline{3\arrayrulewidth}
\end{tabu}
\caption{The impact of bagging on the query selection and on the prediction stages.} \label{tbl:decition-making}
\end{table}

Certain tasks such as general event detection use templates instead of query phrases. Linguistic templates, or extraction patterns \cite{extraction-patt}, cannot be incorporated into our model as is. Future work may explore this direction. Another interesting direction is investigating the efficacy of our model on the platforms with longer documents, e.g., on Amazon or on IMDB. The longer length of documents and the higher occurrence of query phrases can pose exciting research challenges.


%% file: doc-conclusion.tex
In this paper, we proposed an active learning model for social media tasks tailored for query phrases. We employed an algorithm to derive two views from documents, then, we proposed a new multi-view query strategy to aggregate the representativeness and the contention reduction measures. Finally, we proposed an algorithm based on the agreement between multiple predictors to tackle noisy content. Through an extensive set of experiments in four public datasets we showed that our model outperforms existing baselines. We also reported two ablation studies and extensively analyzed our model.

%% file: doc-ethics.tex
\noindent\textbf{Limitations in methodology.} We have proposed an active learning model for a category of tasks called Query-Based problems. We argued in the paper that this category covers a large and diverse set of scenarios. Nonetheless, this category is not a complete set by any means. There exist tasks that fall outside this set and are not query-based, e.g., general offensive language detection and certain event detection tasks. Our model, as is, cannot be used to perform such tasks. To address this limitation our model should be able to incorporate extraction patterns. This enables our model to handle tasks that use lexical templates, such as some event detection problems. We may explore this direction in the future.

\noindent\textbf{Limitations in experiments.} The efficacy of our model depends on the expressiveness of the underlying views extracted from documents. These views are based on contextual word embeddings. Using four English datasets, we experimentally showed that the views extracted by the pretrained BERT are sufficient for this. One can ask whether such views can be extracted from non-English language models. Our paper focuses on English tasks only, and has not explored other languages.

\noindent\textbf{Failure mode and potential misuse.} If our model is used as described in this paper, we expect that it enhances model performance. We used four datasets across various domains to reduce the risk of any bias. Nonetheless, every domain has a specific data distribution and this can affect the efficacy of active learning algorithms \cite{inactive}.

\noindent\textbf{Privacy.} All the datasets used in our paper are public and have been recently used in NLP venues. The publishers of these datasets have targeted important applications that can benefit the society. Nonetheless, if one decides to use our model for processing a new task, they should follow the corresponding terms of service. Most importantly, they should not collect any personal information about users, and should not use our model in sensitive tasks that violate user privacy.

\noindent\textbf{Costs.} Our model aims at reducing the cost of development. Nonetheless, to design our model and to run the baselines, we ran the experiments several dozens of times using a system with a 16-core processor and eight NVIDIA Titan RTX GPUs. Note that in a deployment environment, there is no need to consume such a resource. Because our model can be used as is.

%% file: doc-appendix.tex
\begin{table*}
\centering
\small
\begin{tabu}{p{0.5in}  p{1.35in} p{1.5in} p{1.35in} } \Xhline{3\arrayrulewidth}
\textbf{Dataset} & \textbf{Document Length (Mean)} & \textbf{Document Length (Median)} & \textbf{Total \# of Unique Tokens} \\ \Xhline{3\arrayrulewidth}
\PRODUCTDT & 15.7 & 15 & 14,821 \\ 
\RUMOURDT & 16.1 & 16 & 13,091 \\ 
\ADRDT & 16.4 & 17 & 43,413 \\ 
\OBSERVATIONDT & 14.7 & 15 & 28,398 \\ 
\Xhline{3\arrayrulewidth}
\end{tabu}
\caption{The average and the median of the documents in the datasets (in number of tokens), the third column shows the total number of unique tokens in each dataset. To tokenize the documents we used the parser developed by \citet{tw-parser}. The numbers are without punctuation marks.} \label{tbl:dt-doc-len}
\end{table*}
\begin{table*}
\centering
\small
\begin{tabu}{p{0.5in}  p{0.9in} p{1.2in} p{1.8in} } \Xhline{3\arrayrulewidth}
\textbf{Dataset} & \textbf{\# of Used Queries} & \textbf{Query Length (Mean)} & \textbf{\# of Queries in Document (Mean)} \\ \Xhline{3\arrayrulewidth}
\PRODUCTDT & 2 & 1 & 1.2 \\ 
\RUMOURDT & 6 & 2 & 1 \\ 
\ADRDT & 493 & 1 & 1.3 \\ 
\OBSERVATIONDT & 4 & 1 & 0.95 \\ 
\Xhline{3\arrayrulewidth}
\end{tabu}
\caption{The total number of used queries to collect the datasets, the average length of used queries (in number of words), and the average number of queries appearing in a document.} \label{tbl:dt-query-len}
\end{table*}

\section{Data Analysis} \label{sec:append-data-anal}
In this section, we investigate two relevant aspects of the datasets used in the experiments. Table \ref{tbl:dt-doc-len} reports the average length of the documents in each dataset. We see that the ADR dataset, which is also the largest one, has the largest set of unique tokens. 

Table \ref{tbl:dt-query-len} reports the characteristics of the queries used to collect the datasets. We see that the query set of ADR is extremely large. Note that the average number of queries in a document in the Observation dataset is less than 1. This is despite the fact that the creators of the dataset report that they have collected the data using keywords \cite{crisis-wit}. As a patch for such cases, we used the vector representation of the document view as the vector representation of the keyword view--essentially ignoring the missing keywords.

%% file: acl_latex.bbl
\begin{thebibliography}{61}
\expandafter\ifx\csname natexlab\endcsname\relax\def\natexlab#1{#1}\fi

\bibitem[{Abe and Mamitsuka(1998)}]{qbc-by-bag}
Naoki Abe and Hiroshi Mamitsuka. 1998.
\newblock Query learning strategies using boosting and bagging.
\newblock In \emph{Proc of the 5th ICML}, pages 1--9.

\bibitem[{Abulaish et~al.(2020)Abulaish, Kamal, and Zaki}]{figurative-survey}
Muhammad Abulaish, Ashraf Kamal, and Mohammed~J. Zaki. 2020.
\newblock A survey of figurative language and its computational detection in
  online social networks.
\newblock \emph{{ACM} Transactions on the Web}, 14(1):3:1--3:52.

\bibitem[{Amiri and Daume(2015)}]{churn}
Hadi Amiri and Hal Daume. 2015.
\newblock Target-dependent churn classification in microblogs.
\newblock In \emph{Proceedings of the Twenty-Ninth {AAAI} Conference on
  Artificial Intelligence, January 25-30, 2015, Austin, Texas, {USA}}, pages
  2361--2367. {AAAI} Press.

\bibitem[{An et~al.(2021)An, Kwak, Lee, Jun, and Ahn}]{hate-speech}
Jisun An, Haewoon Kwak, Claire~Seungeun Lee, Bogang Jun, and Yong{-}Yeol Ahn.
  2021.
\newblock Predicting anti-asian hateful users on twitter during {COVID-19}.
\newblock In \emph{Findings of the Association for Computational Linguistics:
  {EMNLP} 2021, Virtual Event / Punta Cana, Dominican Republic, 16-20 November,
  2021}, pages 4655--4666. Association for Computational Linguistics.

\bibitem[{Attenberg and Provost(2011)}]{inactive}
Josh Attenberg and Foster Provost. 2011.
\newblock Inactive learning?: Difficulties employing active learning in
  practice.
\newblock \emph{KDD Exp. News.}, 12:36--41.

\bibitem[{Balcan et~al.(2005)Balcan, Blum, and Yang}]{cotrain-ortho}
Maria-Florina Balcan, Avrim Blum, and Ke~Yang. 2005.
\newblock Co-training and expansion: Towards bridging theory and practice.
\newblock In \emph{NIPS}, pages 89--96.

\bibitem[{Bevilacqua et~al.(2021)Bevilacqua, Pasini, Raganato, and
  Navigli}]{embed-wsd-survey}
Michele Bevilacqua, Tommaso Pasini, Alessandro Raganato, and Roberto Navigli.
  2021.
\newblock Recent trends in word sense disambiguation: {A} survey.
\newblock In \emph{Proceedings of the Thirtieth International Joint Conference
  on Artificial Intelligence, {IJCAI} 2021, Virtual Event / Montreal, Canada,
  19-27 August 2021}, pages 4330--4338. ijcai.org.

\bibitem[{Biddle et~al.(2020)Biddle, Joshi, Liu, Paris, and Xu}]{phm-new}
Rhys Biddle, Aditya Joshi, Shaowu Liu, Cecile Paris, and Guandong Xu. 2020.
\newblock Leveraging sentiment distributions to distinguish figurative from
  literal health reports on twitter.
\newblock In \emph{Proceedings of The Web Conference 2020}, WWW ’20, page
  1217–1227, New York, NY, USA. Association for Computing Machinery.

\bibitem[{Buhlmann and Yu(2002)}]{bagging-analysis}
Peter Buhlmann and Bin Yu. 2002.
\newblock Analyzing bagging.
\newblock \emph{Ann. Statist.}, 30(4):927--961.

\bibitem[{Burkhardt et~al.(2020)Burkhardt, Siekiera, Glodde, Andrade-Navarro,
  and Kramer}]{active-adr-2}
Sophie Burkhardt, Julia Siekiera, Josua Glodde, Miguel~A Andrade-Navarro, and
  Stefan Kramer. 2020.
\newblock Towards identifying drug side effects from social media using active
  learning and crowd sourcing.
\newblock In \emph{Pacific Symposium of Biocomputing (PSB)}, pages 319--330.

\bibitem[{Chapelle et~al.(2003)Chapelle, Weston, and
  Sch\"{o}lkopf}]{cluster-kernel}
Olivier Chapelle, Jason Weston, and Bernhard Sch\"{o}lkopf. 2003.
\newblock Cluster kernels for semi-supervised learning.
\newblock In \emph{NIPS 15}, pages 601--608. MIT Press.

\bibitem[{Devlin et~al.(2019)Devlin, Chang, Lee, and Toutanova}]{bert}
Jacob Devlin, Ming-Wei Chang, Kenton Lee, and Kristina Toutanova. 2019.
\newblock {BERT}: Pre-training of deep bidirectional transformers for language
  understanding.
\newblock In \emph{Proc of the 2019 NAACL}, pages 4171--4186.

\bibitem[{Efron(2011)}]{microblogs}
Miles Efron. 2011.
\newblock Information search and retrieval in microblogs.
\newblock \emph{Journal of the American Society for Information Science and
  Technology (JASIST)}, 62(6):996--1008.

\bibitem[{Ein{-}Dor et~al.(2020)Ein{-}Dor, Halfon, Gera, Shnarch, Dankin,
  Choshen, Danilevsky, Aharonov, Katz, and Slonim}]{bert-al}
Liat Ein{-}Dor, Alon Halfon, Ariel Gera, Eyal Shnarch, Lena Dankin, Leshem
  Choshen, Marina Danilevsky, Ranit Aharonov, Yoav Katz, and Noam Slonim. 2020.
\newblock Active learning for {BERT:} an empirical study.
\newblock In \emph{Proceedings of the 2020 Conference on Empirical Methods in
  Natural Language Processing, {EMNLP} 2020, Online, November 16-20, 2020},
  pages 7949--7962.

\bibitem[{Farinneya et~al.(2021)Farinneya, Pour, Hamidian, and
  Diab}]{al-rumour}
Parsa Farinneya, Mohammad Mahdi~Abdollah Pour, Sardar Hamidian, and Mona~T.
  Diab. 2021.
\newblock Active learning for rumor identification on social media.
\newblock In \emph{Findings of the Association for Computational Linguistics:
  {EMNLP} 2021, Virtual Event / Punta Cana, Dominican Republic, 16-20 November,
  2021}, pages 4556--4565.

\bibitem[{Farzindar and Inkpen(2020)}]{nlp-soc-med}
Anna~Atefeh Farzindar and Diana Inkpen. 2020.
\newblock \emph{Natural Language Processing for Social Media, Third Edition}.
\newblock Synthesis Lectures on Human Language Technologies. Morgan {\&}
  Claypool Publishers.

\bibitem[{Gareth et~al.(2013)Gareth, Daniela, Trevor, and
  Robert}]{intro-ml-book}
James Gareth, Witten Daniela, Hastie Trevor, and Tibshirani Robert. 2013.
\newblock \emph{An introduction to statistical learning: with applications in
  R}.
\newblock Spinger.

\bibitem[{Ghani et~al.(2003)Ghani, Jones, Mitchell, and Riloff}]{co-testing-2}
Rayid Ghani, Rosie Jones, Tom Mitchell, and Ellen Riloff. 2003.
\newblock Active learning for information extraction with multiple view feature
  sets.
\newblock In \emph{Proc of the 20th ICML}, pages 26--34.

\bibitem[{Guo et~al.(2017)Guo, Pleiss, Sun, and Weinberger}]{cls-conf}
Chuan Guo, Geoff Pleiss, Yu~Sun, and Kilian~Q. Weinberger. 2017.
\newblock On calibration of modern neural networks.
\newblock In \emph{Proceedings of the 34th International Conference on Machine
  Learning, {ICML} 2017, Sydney, NSW, Australia, 6-11 August 2017}, volume~70
  of \emph{Proceedings of Machine Learning Research}, pages 1321--1330.

\bibitem[{Heidenreich et~al.(2013)Heidenreich, Schindler, and
  Sperlich}]{kernel-de}
Nils-Bastian Heidenreich, Anja Schindler, and Stefan Sperlich. 2013.
\newblock Bandwidth selection for kernel density estimation: a review of fully
  automatic selectors.
\newblock \emph{AStA Advances in Statistical Analysis}, 97(4):403--433.

\bibitem[{Imran et~al.(2015)Imran, Castillo, Diaz, and Vieweg}]{crisis-survey}
Muhammad Imran, Carlos Castillo, Fernando Diaz, and Sarah Vieweg. 2015.
\newblock Processing social media messages in mass emergency: A survey.
\newblock \emph{ACM Comput. Surv.}, 47(4):67:1--67:38.

\bibitem[{Jedoui et~al.(2019)Jedoui, Krishna, Bernstein, and
  Fei-Fei}]{active-multi-ans}
Khaled Jedoui, Ranjay Krishna, Michael Bernstein, and Li~Fei-Fei. 2019.
\newblock Deep bayesian active learning for multiple correct outputs.
\newblock \emph{arXiv preprint arXiv:1912.01119}.

\bibitem[{Jiang et~al.(2020)Jiang, Gao, Duan, Kang, Sun, Zhang, and
  Liu}]{act-spam}
Zhuoren Jiang, Zhe Gao, Yu~Duan, Yangyang Kang, Changlong Sun, Qiong Zhang, and
  Xiaozhong Liu. 2020.
\newblock Camouflaged {C}hinese spam content detection with semi-supervised
  generative active learning.
\newblock In \emph{Proceedings of the 58th Annual Meeting of the Association
  for Computational Linguistics}, pages 3080--3085, Online. Association for
  Computational Linguistics.

\bibitem[{Karisani and Karisani(2020)}]{our-corona}
Negin Karisani and Payam Karisani. 2020.
\newblock \href {https://arxiv.org/abs/2004.06778} {Mining coronavirus
  (covid-19) posts in social media}.
\newblock \emph{arXiv preprint arXiv:2004.06778}.

\bibitem[{Karisani and Agichtein(2018)}]{wespad}
Payam Karisani and Eugene Agichtein. 2018.
\newblock Did you just have a heart attack?: Towards robust detection of
  personal health mentions in social media.
\newblock In \emph{Proc of the 2018 WWW}, pages 137--146.

\bibitem[{Karisani et~al.(2020)Karisani, Agichtein, and Ho}]{co-decomp}
Payam Karisani, Eugene Agichtein, and Joyce Ho. 2020.
\newblock Domain-guided task decomposition with self-training for detecting
  personal events in social media.
\newblock In \emph{Proceedings of The Web Conference 2020}, WWW ’20, page
  2411–2420, New York, NY, USA. Association for Computing Machinery.

\bibitem[{Karisani et~al.(2021)Karisani, Choi, and Xiong}]{view-distill}
Payam Karisani, Jinho~D. Choi, and Li~Xiong. 2021.
\newblock \href {http://arxiv.org/abs/2105.11354} {View distillation with
  unlabeled data for extracting adverse drug effects from user-generated data}.

\bibitem[{Karisani and Karisani(2021)}]{self-pretraining}
Payam Karisani and Negin Karisani. 2021.
\newblock Semi-supervised text classification via self-pretraining.
\newblock In \emph{Proceedings of the 14th ACM International Conference on Web
  Search and Data Mining}, WSDM '21, page 40–48. Association for Computing
  Machinery.

\bibitem[{Kochkina et~al.(2018)Kochkina, Liakata, and Zubiaga}]{rumour-dt}
Elena Kochkina, Maria Liakata, and Arkaitz Zubiaga. 2018.
\newblock All-in-one: Multi-task learning for rumour verification.
\newblock In \emph{Proceedings of the 27th International Conference on
  Computational Linguistics, {COLING} 2018, Santa Fe, New Mexico, USA, August
  20-26, 2018}, pages 3402--3413. Association for Computational Linguistics.

\bibitem[{Kong et~al.(2014)Kong, Schneider, Swayamdipta, Bhatia, Dyer, and
  Smith}]{tw-parser}
Lingpeng Kong, Nathan Schneider, Swabha Swayamdipta, Archna Bhatia, Chris Dyer,
  and Noah~A. Smith. 2014.
\newblock A dependency parser for tweets.
\newblock In \emph{Proceedings of the 2014 Conference on Empirical Methods in
  Natural Language Processing, ({EMNLP} 2014)}, pages 1001--1012.

\bibitem[{Konyushkova et~al.(2017)Konyushkova, Sznitman, and Fua}]{lal}
Ksenia Konyushkova, Raphael Sznitman, and Pascal Fua. 2017.
\newblock Learning active learning from data.
\newblock In \emph{Advances in Neural Information Processing Systems (NIPS)
  30}, pages 4225--4235. Curran Associates, Inc.

\bibitem[{Kullback and Leibler(1951)}]{kl-div}
Solomon Kullback and Richard~A Leibler. 1951.
\newblock On information and sufficiency.
\newblock \emph{The annals of mathematical statistics}, 22(1):79--86.

\bibitem[{Lewis and Gale(1994)}]{uncertainty}
David~D. Lewis and William~A. Gale. 1994.
\newblock A sequential algorithm for training text classifiers.
\newblock In \emph{Proc of the 17th SIGIR}, pages 3--12.

\bibitem[{Liao and Grishman(2011)}]{sent-cotest}
Shasha Liao and Ralph Grishman. 2011.
\newblock Using prediction from sentential scope to build a pseudo co-testing
  learner for event extraction.
\newblock In \emph{Proc of 5th IJCNLP}, pages 714--722.

\bibitem[{Lowell et~al.(2019)Lowell, Lipton, and Wallace}]{active-practical}
David Lowell, Zachary~C. Lipton, and Byron~C. Wallace. 2019.
\newblock Practical obstacles to deploying active learning.
\newblock In \emph{Proc of the 2019 EMNLP}, pages 21--30.

\bibitem[{Magge et~al.(2021)Magge, Klein, Miranda-Escalada, Ali Al-Garadi,
  Alimova, Miftahutdinov, Farre, Lima~L{\'o}pez, Flores, O{'}Connor,
  Weissenbacher, Tutubalina, Sarker, Banda, Krallinger, and
  Gonzalez-Hernandez}]{smm4h-2021}
Arjun Magge, Ari Klein, Antonio Miranda-Escalada, Mohammed Ali Al-Garadi,
  Ilseyar Alimova, Zulfat Miftahutdinov, Eulalia Farre, Salvador
  Lima~L{\'o}pez, Ivan Flores, Karen O{'}Connor, Davy Weissenbacher, Elena
  Tutubalina, Abeed Sarker, Juan Banda, Martin Krallinger, and Graciela
  Gonzalez-Hernandez. 2021.
\newblock Overview of the sixth social media mining for health applications
  ({\#}{SMM}4{H}) shared tasks at {NAACL} 2021.
\newblock In \emph{Proceedings of the Sixth Social Media Mining for Health
  ({\#}SMM4H) Workshop and Shared Task}, Mexico City, Mexico. Association for
  Computational Linguistics.

\bibitem[{Margatina et~al.(2021)Margatina, Vernikos, Barrault, and
  Aletras}]{cal}
Katerina Margatina, Giorgos Vernikos, Lo{\"{\i}}c Barrault, and Nikolaos
  Aletras. 2021.
\newblock Active learning by acquiring contrastive examples.
\newblock In \emph{Proceedings of the 2021 Conference on Empirical Methods in
  Natural Language Processing, {EMNLP} 2021, Virtual Event / Punta Cana,
  Dominican Republic, 7-11 November, 2021}, pages 650--663.

\bibitem[{Mccreadie et~al.(2019)Mccreadie, Buntain, and Soboroff}]{perf-metric}
R.~Mccreadie, C.~Buntain, and I.~Soboroff. 2019.
\newblock Trec incident streams: Actionable information on social media.
\newblock In \emph{Proc of the 16th ISCRAM}.

\bibitem[{Muslea et~al.(2000)Muslea, Minton, and Knoblock}]{co-testing-0}
Ion Muslea, Steven Minton, and Craig~A. Knoblock. 2000.
\newblock Selective sampling with redundant views.
\newblock In \emph{Proc of the Seventeenth National Conference on Artificial
  Intelligence and Twelfth Conference on Innovative Applications of Artificial
  Intelligence}, pages 621--626. AAAI Press.

\bibitem[{Nguyen and Smeulders(2004)}]{density-clustering}
Hieu~T. Nguyen and Arnold Smeulders. 2004.
\newblock Active learning using pre-clustering.
\newblock In \emph{Proc of the Twenty-first International Conference on Machine
  Learning}, ICML '04, pages 79--.

\bibitem[{Nigam and Ghani(2000)}]{cotrain-2}
Kamal Nigam and Rayid Ghani. 2000.
\newblock Analyzing the effectiveness and applicability of co-training.
\newblock In \emph{Proc of the 2000 {ACM} {CIKM} International Conference on
  Information and Knowledge Management, McLean, VA, USA, November 6-11, 2000},
  pages 86--93.

\bibitem[{Riloff and Jones(1999)}]{extraction-patt}
Ellen Riloff and Rosie Jones. 1999.
\newblock Learning dictionaries for information extraction by multi-level
  bootstrapping.
\newblock In \emph{Proceedings of the Sixteenth National Conference on
  Artificial Intelligence and Eleventh Conference on Innovative Applications of
  Artificial Intelligence, July 18-22, 1999, Orlando, Florida, {USA}}, pages
  474--479. {AAAI} Press / The {MIT} Press.

\bibitem[{Ru et~al.(2020)Ru, Feng, Qiu, Zhou, Wang, Zhang, Yu, and
  Li}]{al-adv-uncertain}
Dongyu Ru, Jiangtao Feng, Lin Qiu, Hao Zhou, Mingxuan Wang, Weinan Zhang, Yong
  Yu, and Lei Li. 2020.
\newblock Active sentence learning by adversarial uncertainty sampling in
  discrete space.
\newblock In \emph{Findings of the Association for Computational Linguistics:
  {EMNLP} 2020, Online Event, 16-20 November 2020}, volume {EMNLP} 2020 of
  \emph{Findings of {ACL}}, pages 4908--4917.

\bibitem[{Sawhney et~al.(2020)Sawhney, Joshi, Gandhi, and Shah}]{suicide}
Ramit Sawhney, Harshit Joshi, Saumya Gandhi, and Rajiv~Ratn Shah. 2020.
\newblock A time-aware transformer based model for suicide ideation detection
  on social media.
\newblock In \emph{Proceedings of the 2020 Conference on Empirical Methods in
  Natural Language Processing, {EMNLP} 2020, Online, November 16-20, 2020},
  pages 7685--7697. Association for Computational Linguistics.

\bibitem[{Settles(2009)}]{act-ler-survey}
Burr Settles. 2009.
\newblock Active learning literature survey.
\newblock Computer Sciences Technical Report 1648, University of
  Wisconsin--Madison.

\bibitem[{Shi and Lin(2019)}]{bert-srl}
Peng Shi and Jimmy Lin. 2019.
\newblock Simple bert models for extraction and semantic role labeling.
\newblock \emph{arXiv preprint arXiv:1904.05255}.

\bibitem[{Siddhant and Lipton(2018)}]{nlp-al}
Aditya Siddhant and Zachary~C. Lipton. 2018.
\newblock Deep bayesian active learning for natural language processing:
  Results of a large-scale empirical study.
\newblock In \emph{Proceedings of the 2018 Conference on Empirical Methods in
  Natural Language Processing, Brussels, Belgium, October 31 - November 4,
  2018}, pages 2904--2909. Association for Computational Linguistics.

\bibitem[{Silverman(1986)}]{kde-est}
Bernard~W Silverman. 1986.
\newblock \emph{Density estimation for statistics and data analysis},
  volume~26.
\newblock CRC press.

\bibitem[{Stanovsky et~al.(2017)Stanovsky, Gruhl, and Mendes}]{active-adr}
Gabriel Stanovsky, Daniel Gruhl, and P~Mendes. 2017.
\newblock Recognizing mentions of adverse drug reaction in social media using
  knowledge-infused recurrent models.
\newblock In \emph{Proc of the 15th EACL}, pages 142--151.

\bibitem[{Sun(2013)}]{multi-view-published}
Shiliang Sun. 2013.
\newblock A survey of multi-view machine learning.
\newblock \emph{Neural Computing and Applications}, 23(7-8):2031--2038.

\bibitem[{Swayamdipta et~al.(2020)Swayamdipta, Schwartz, Lourie, Wang,
  Hajishirzi, Smith, and Choi}]{data-maps}
Swabha Swayamdipta, Roy Schwartz, Nicholas Lourie, Yizhong Wang, Hannaneh
  Hajishirzi, Noah~A. Smith, and Yejin Choi. 2020.
\newblock Dataset cartography: Mapping and diagnosing datasets with training
  dynamics.
\newblock In \emph{Proceedings of the 2020 Conference on Empirical Methods in
  Natural Language Processing, {EMNLP} 2020, Online, November 16-20, 2020},
  pages 9275--9293. ACL.

\bibitem[{Weissenbacher et~al.(2019)Weissenbacher, Sarker, Magge, Daughton,
  O{'}Connor, Paul, and Gonzalez-Hernandez}]{smm4h}
Davy Weissenbacher, Abeed Sarker, Arjun Magge, Ashlynn Daughton, Karen
  O{'}Connor, Michael~J. Paul, and Graciela Gonzalez-Hernandez. 2019.
\newblock \href {https://doi.org/10.18653/v1/W19-3203} {Overview of the fourth
  social media mining for health ({SMM}4{H}) shared tasks at {ACL} 2019}.
\newblock In \emph{Proceedings of the Fourth Social Media Mining for Health
  Applications ({\#}SMM4H) Workshop {\&} Shared Task}, pages 21--30, Florence,
  Italy. Association for Computational Linguistics.

\bibitem[{Wolf et~al.(2019)Wolf, Debut, and et~al.}]{bert-impl}
Thomas Wolf, Lysandre Debut, and et~al. 2019.
\newblock Huggingface's transformers: State-of-the-art natural language
  processing.
\newblock \emph{ArXiv}, abs/1910.03771.

\bibitem[{Wright and Augenstein(2020)}]{dom-ada-mixture-bert}
Dustin Wright and Isabelle Augenstein. 2020.
\newblock Transformer based multi-source domain adaptation.
\newblock In \emph{Proceedings of the 2020 Conference on Empirical Methods in
  Natural Language Processing, {EMNLP} 2020, Online, November 16-20, 2020},
  pages 7963--7974. Association for Computational Linguistics.

\bibitem[{Xu et~al.(2013)Xu, Tao, and Xu}]{multi-view}
Chang Xu, Dacheng Tao, and Chao Xu. 2013.
\newblock \href {http://arxiv.org/abs/1304.5634} {A survey on multi-view
  learning}.
\newblock \emph{CoRR}, abs/1304.5634.

\bibitem[{Xu et~al.(2021)Xu, Ritter, Baldwin, and Rahimi}]{wnut-2021}
Wei Xu, Alan Ritter, Tim Baldwin, and Afshin Rahimi, editors. 2021.
\newblock \emph{Proceedings of the Seventh Workshop on Noisy User-generated
  Text (W-NUT 2021) co-located with the Conference on Empirical Methods in
  Natural Language Processing (EMNLP)}. Association for Computational
  Linguistics, Online.

\bibitem[{Yousef et~al.(2021)Yousef, Schlaf, Borst, Niekler, and
  Heyer}]{press-free}
Tariq Yousef, Antje Schlaf, Janos Borst, Andreas Niekler, and Gerhard Heyer.
  2021.
\newblock Press freedom monitor: Detection of reported press and media freedom
  violations in twitter and news articles.
\newblock In \emph{Proceedings of the 2021 Conference on Empirical Methods in
  Natural Language Processing: System Demonstrations, {EMNLP} 2021, Online and
  Punta Cana, Dominican Republic, 7-11 November, 2021}, pages 153--159.
  Association for Computational Linguistics.

\bibitem[{Yuan et~al.(2020)Yuan, Lin, and Boyd{-}Graber}]{alps}
Michelle Yuan, Hsuan{-}Tien Lin, and Jordan~L. Boyd{-}Graber. 2020.
\newblock Cold-start active learning through self-supervised language modeling.
\newblock In \emph{Proceedings of the 2020 Conference on Empirical Methods in
  Natural Language Processing, {EMNLP} 2020, Online, November 16-20, 2020},
  pages 7935--7948. Association for Computational Linguistics.

\bibitem[{Zahra et~al.(2020)Zahra, Imran, and Ostermann}]{crisis-wit}
Kiran Zahra, Muhammad Imran, and Frank~O. Ostermann. 2020.
\newblock Automatic identification of eyewitness messages on twitter during
  disasters.
\newblock \emph{Information Processing \& Management}, 57(1):102107.

\bibitem[{Zhang and Plank(2021)}]{caral}
Mike Zhang and Barbara Plank. 2021.
\newblock Cartography active learning.
\newblock In \emph{Proceedings of the 2021 Conference on Empirical Methods in
  Natural Language Processing, {EMNLP} 2021, Virtual Event / Punta Cana,
  Dominican Republic, 7-11 November, 2021}, pages 395--406.

\bibitem[{Zhao et~al.(2020)Zhao, Prosperi, Lyu, Guo, and Bian}]{act-job-tweet}
Yunpeng Zhao, Mattia Prosperi, Tianchen Lyu, Yi~Guo, and Jing Bian. 2020.
\newblock Integrating crowdsourcing and active learning for classification of
  work-life events from tweets.
\newblock \emph{arXiv preprint arXiv:2003.12139}.

\end{thebibliography}
